\documentclass[conference]{IEEEtran}
\usepackage{graphicx}
\usepackage{amsmath}
\usepackage{caption}

\title{\LARGE \bf
SAVMap: Structure-Aided Visual Mapping of Large-Scale 2.5D Manhattan Wireframes from Panoramic Video
}

\author{
  \IEEEauthorblockN{Howard Huang, Bharath Surianarayanan, Keifer Lee, Chenyu Wang, Chen Feng}
  \thanks{All authors were affiliated with Nokia Bell Labs, except Chen Feng who is with NYU. howard.huang@ieee.org, bs4224@nyu.edu, kl3866@nyu.edu, wangchy4@shanghaitech.edu.cn, cfeng@nyu.edu}
}

\let\oldtwocolumn\twocolumn
\renewcommand\twocolumn[1][]{%
    \oldtwocolumn[{#1}{
    \begin{center}
        \vspace{-6mm}\includegraphics[trim=0 240 0 175, clip, width=1.1\linewidth]{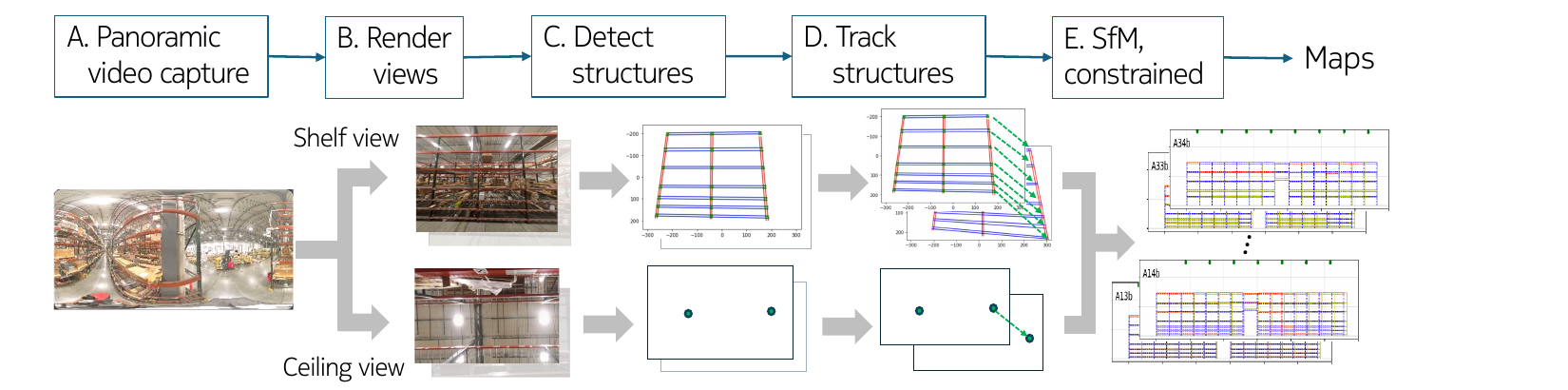}
        \captionof{figure}{SAVMap generates per-aisle wireframe maps of warehouse lights and shelves from a panoramic video.}
        \label{fig:overview}
        \end{center}
    }]
}

\begin{document}

\maketitle
\thispagestyle{empty}
\pagestyle{empty}

\begin{abstract}
Precise 3D representations of industrial environments enable tasks such as robot localization and digital twin generation. We propose SAVMap, a method for generating a semantic wireframe map of warehouse shelf and light structures using only a panoramic video camera as the sensor input. Sequences of rectified images with shelf and ceiling-facing views are extracted from a panoramic video captured along the warehouse aisles. Using a semantic segmentation network front end, a set of sparse, semantic structure feature points (e.g., corners of shelf structures, centers of lights) are extracted from each image and tracked across the sequences. By accounting for real-world geometric relationships among the points such as Manhattan grids, a constrained structure-from-motion algorithm yields the 3D points that form a wireframe map. We demonstrate the scalability and accuracy of our proposal in a warehouse with 46 shelving rows, each with faces spanning 55\,m by 7\,m. From an hour of panoramic video content, we create wireframe maps for over 5000 shelf elements across the rows, achieving an aggregate mean absolute error of 4.8\,cm with respect to ground-truth.  
\end{abstract}

\section{Introduction}
\label{sec:introduction}
Digital 3D representations of structured environments such as warehouses and factories enable important foundational technologies such as robot localization and digital twin generation. These technologies can in turn be used for higher-level applications such as collision avoidance, process optimization, and inventory management. 

Point clouds are often used as 3D map representations because of their versatility and accuracy, and there is a rich literature on their generation using various sensor modalities (e.g., laser scanner, RGB, RGB-D cameras) and applying both classical and learning-based strategies. While effective for general-purpose mapping, point clouds provide no semantic information, and when derived from 2D images, visually repetitive environments can lead to drift, distortion, or outright failure.

\begin{figure}
    \centering
    \includegraphics[trim=109 165 0 105, clip, width=1.45\linewidth]{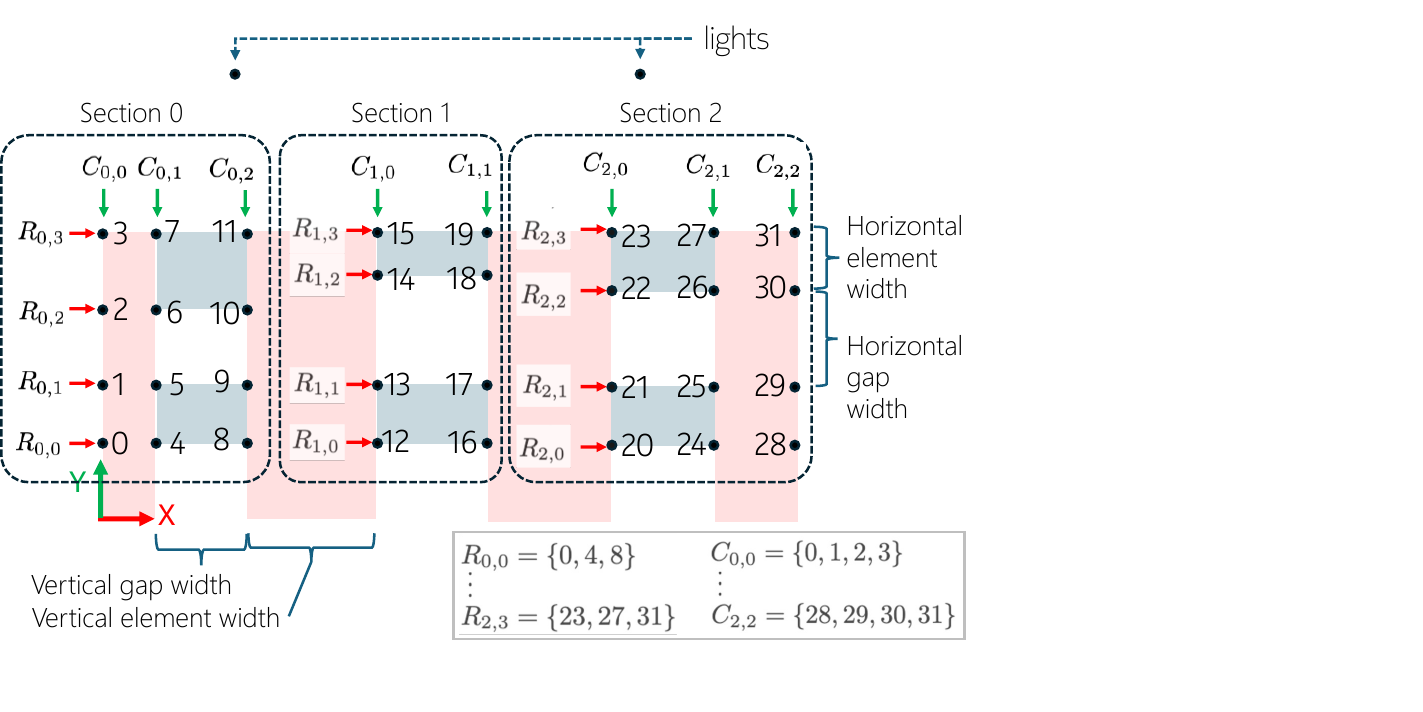}
    \caption{The SAVMap wireframe map consists of connected feature points, each with a 3D coordinate and semantic label. The numbers indicate point labels used in the SfM optimization, and sets $C_{s,i}$ and $R_{s,i}$ contain points for column/row $i$ of section $s$}
    \label{fig:intersection}
\end{figure}

To address these challenges, we propose Structure-Aided Visual Mapping (SAVMap), a pipeline that constructs a sparse, semantically meaningful wireframe of feature points based on structures in the environment. For example, the warehouse shelving face in Figure \ref{fig:intersection} consists horizontal and vertical elements, and the feature points are the edge vertices. SAVMap estimates the coordinates of all edge vertices in an aisle by imposing geometric constraints such as the Manhattan grid. From the 3D coordinates and semantic labels (e.g., “feature 15 is the upper-left corner of the top horizontal element of Section 1”) the spatial occupancy of the shelf face can be characterized.  

We note that, if necessary, the wireframe map could be aligned with a point cloud created via another modality, providing refinement and semantic context. While the shelf face points lie on a 2D plane, they occupy a 3D space when observed by a camera sensor. In this sense, the points are "2.5D'', but in general we will refer to them as 3D. 

The general goal of SAVMap is to accurately reconstruct the labels and 3D coordinates of relevant information-bearing structure points, using only panoramic video input. Our specific environment of interest is a warehouse with repetitive and cluttered shelves. We aim to create a map of only the static shelving and light infrastructure, because they can be relied on for localization if the shelf contents change. We illustrate the end-to-end SAVMap process in Figure \ref{fig:overview}:
\begin{itemize}
\item {\bf Panoramic video capture and view rendering.} Using only a consumer-grade panoramic video camera, SAVMap automatically generates rectified sequences of views (e.g., shelf- or ceiling-facing). The process is simpler than using multiple cameras because it eliminates the need for manual orientation, manual calibration, and camera synchronization. 
\item{\bf Structure detection.}  A lightweight semantic segmentation network identifies structures of interest with binary pixel masks. From the masks, we determine boundary lines whose intersections yield a set of sparse structure-based feature points. Unlike generic visual point detectors (e.g., SIFT, ORB, SuperPoint), this step ensures that points directly correspond to semantically meaningful objects. 
\item{\bf Robust structure tracking.} Correspondences for the structure points need to be established across the image sequences, and we introduce a tracking algorithm to handle missed or spurious detections. 
\item{\bf Constrained Structure from Motion (SfM).} To estimate the 3D coordinates, we propose a modified SfM that incorporates real-world geometric constraints tied to the Manhattan grid.  
\end{itemize}
Using a panoramic video sensor, efficient front-end segmentation targeted on only structures of interest, and light-weight processing of sparse point sets, our proposal is a {\em low cost}, {\em low complexity} pipeline that creates {\em semantic} and {\em accurate} warehouse maps at scale that are {\em robust} against visual clutter.
Our main contributions are as follows:
\begin{enumerate}
\item {\bf New mapping paradigm.} We introduce a framework that leverages semantic understanding with real-world geometric constraints for creating accurate, sparse 3D representations of structured environments from 2D images. 
\item {\bf Robust implementation.} We implement an end-to-end pipeline for this framework, starting from panoramic video. The pipeline extends known techniques and adds handcrafted components to ensure robustness for missing geometric structure. 
\item {\bf Real-world demonstration.} We test our proposal in a real warehouse environment and demonstrate centimeter-level accuracy at scale in comparison to a ground truth map.  
\end{enumerate}

\section{Related Work}
\label{sec:prior_techniques}
\subsection{Direct scanning}
Companies like Leica, Navvis, and Matterport employ laser scanners augmented by RGB cameras to directly create very accurate and dense color point clouds. Contemporary models incorporate IMUs and VIO to enable mobile scanning. While these devices can achieve sub-centimeter accuracy at distances of 10s of meters \cite{rtc360}, they are very expensive and create unordered point clouds with no semantic meaning. (As an exception, Matterport is able to automatically create floor plans \cite{matterport}.) 
In comparison, SAVMap creates minimally sparse yet meaningful maps with much lower device cost. 

\subsection{General 3D reconstruction from 2D images}
Using sets 2D image data, this class of techniques similarly creates unstructured point clouds but with lower density. A baseline implementation of classical techniques like structure from motion and multiview stereo are implemented in the COLMAP software library \cite{colmap_sfm}\cite{colmap_mvs}.  More recently, DUSt3R \cite{dust3r} was introduced as a pioneering transformer-based learning technique for 3D reconstruction. But its performance has been surpassed by MASt3R \cite{mast3r} and then VGGT \cite{wang2025vggt}.  

The classical image-based techniques often struggle in visually repetitive environments in attempting loop closure and bundle adjustment. And as we will see in Section \ref{sec:results}, even VGGT is not robust in these cases either.

\subsection{3D wireframe reconstruction}
The generation of 3D wireframes from 3D point clouds has been addressed by a number of learning-based approaches recently \cite{liu2021pc2wf} \cite{matveev2021scalar}\cite{guo2024wireframe} \cite{ma2024generating}\cite{nonmanhat}, but they require high-quality point clouds and are therefore not relevant for comparison with our approach. 

Classical approaches for 3D line-based representation include generalized SfM frameworks for line segments \cite{BartoliSturm2005}\cite{hofer2013incremental}, a technique for create 3D line clouds (Line3D++) \cite{line3d}, and a generalized library for 3D line mapping \cite{Liu_2023_CVPR} 
Recent learning-based approaches include LC2WF \cite{lc2wf}, SOLD$^2$ \cite{2021sold2}, the winner of a recent 3D reconstruction competition \cite{s23dr}, and Edge-NeRF \cite{edgenerf}. 
As with the point-based approaches, these techniques perform poorly in repetitive environments. SOLD$^2$ may be more robust, but their model is more computationally complex than our lightweight front-end segmentation model. And all other learning-based approaches rely on information (e.g., line clouds, pose information) derived from classical techniques. 



\section{Methodology}
\label{sec:proposed_process}
In this section we describe the detailed steps of our proposal in Figure \ref{fig:overview} for a warehouse environment.

\subsection{Panoramic video capture} 
\begin{figure}
    \centering
    \includegraphics[trim=50 160 0 120, clip, width=1.6\linewidth]{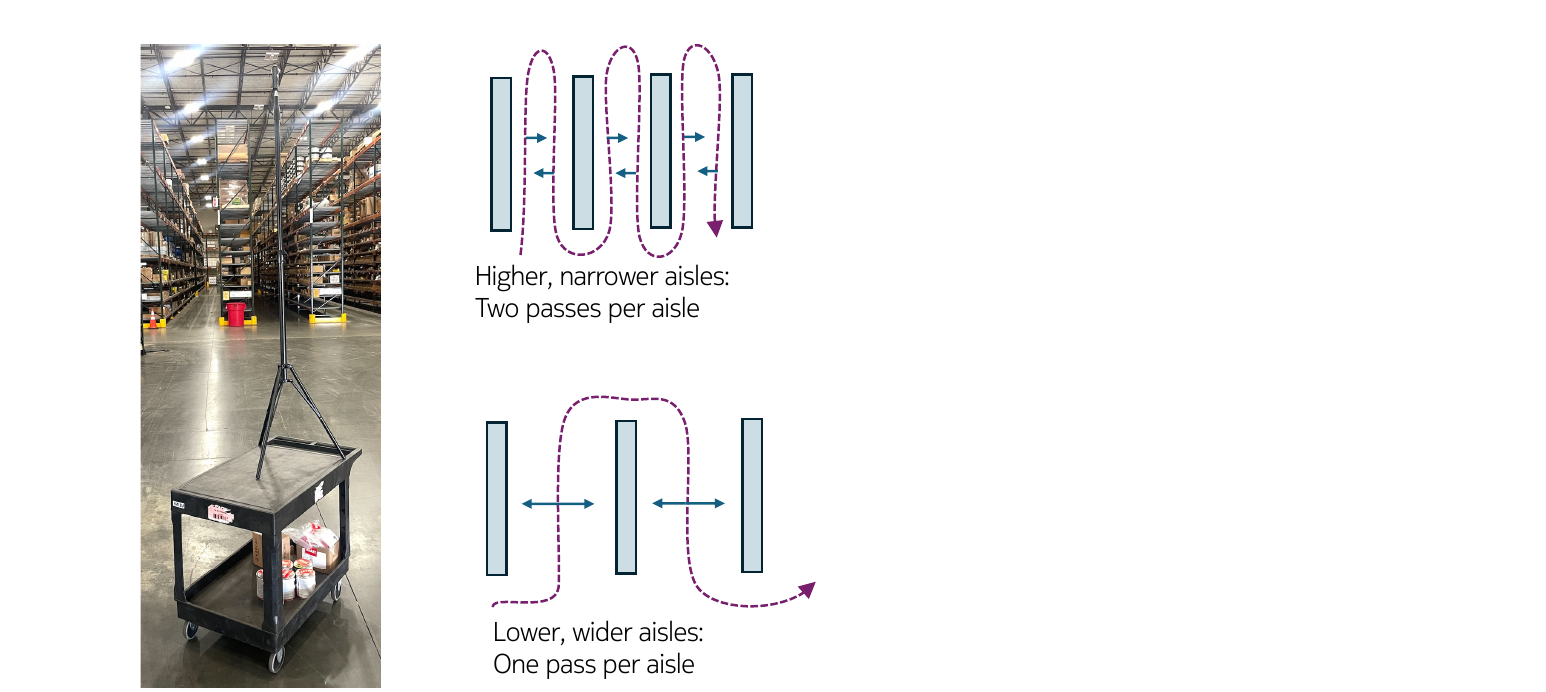}
    \caption{Left: The video capture system (Step A) consists of a panoramic camera mounted on a tripod and cart. Right: Trajectory of cart weaving through the warehouse aisles.
    }
    \label{fig:image_capture}
\end{figure}

The panoramic camera is mounted on a tripod and placed on a cart, with the camera height adjusted so it is roughly centered on the scene of interest. The cart travels up and down the aisles of the warehouse as shown in Figure \ref{fig:image_capture}, possibly requiring two passes per aisle in order for the camera to fully capture the shelf face.  Compared with using multiple synchronized conventional cameras, the panoramic camera offers more flexibility, especially with the automated rendering process described below. 

\subsection{Rendering views.}
For a given aisle, rectilinear shelf-facing and ceiling-facing views are rendered from the panoramic video. While the field of view and camera orientation can be adjusted manually, we develop an automated rendering process (using the opensource py360convert library that supports conversion between panoramic equirectangular and planar formats) which iteratively adjusts the orientation and field of view parameters based on the desired visibility of structures. For the shelf view, the complete vertical extent should be visible for the entire aisle video with minimal a field of view, and for the ceiling view, at least two lights should be visible to resolve yaw estimation ambiguity. 

If the panoramic camera is properly calibrated \cite{aghayari2017}, images rendered from given desired intrinsic parameters will conform to a pinhole model.

\subsection{Detect structures}
\begin{figure}
    \centering
    \includegraphics[trim=140 100 0 90, clip, width=1.3\linewidth]{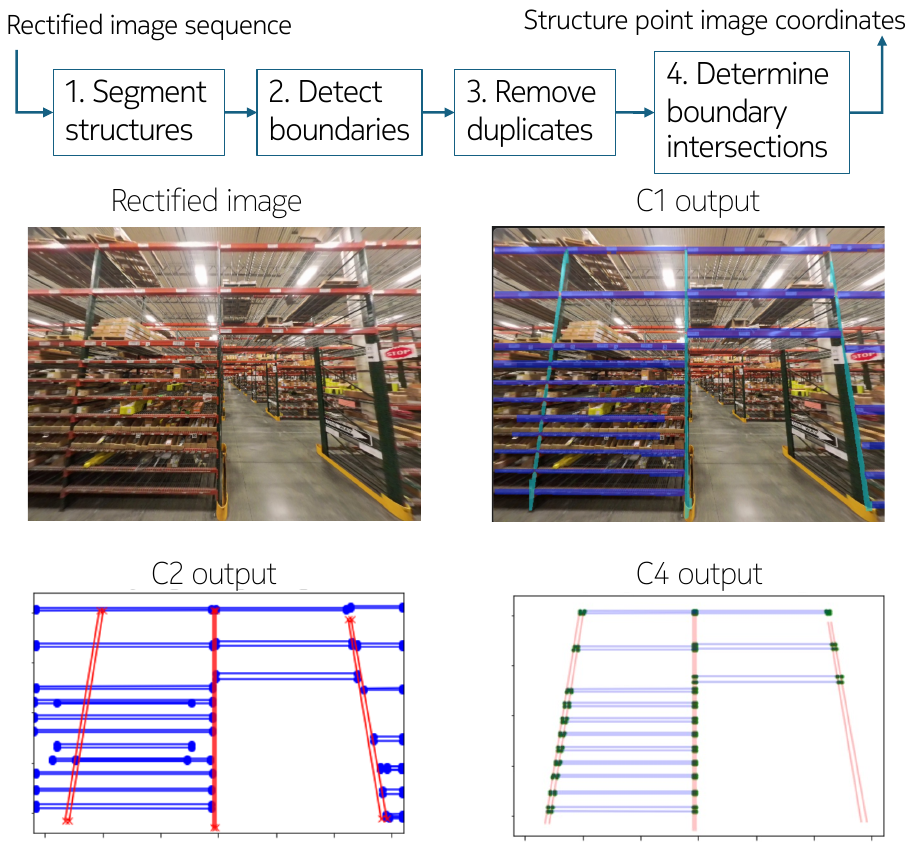}
    \caption{The structure detection process (Step C) and sample data}
    \label{fig:detection}
\end{figure}

\begin{figure}
    \centering
    \includegraphics[trim=180 220 0 190, clip, width=1.02\linewidth]{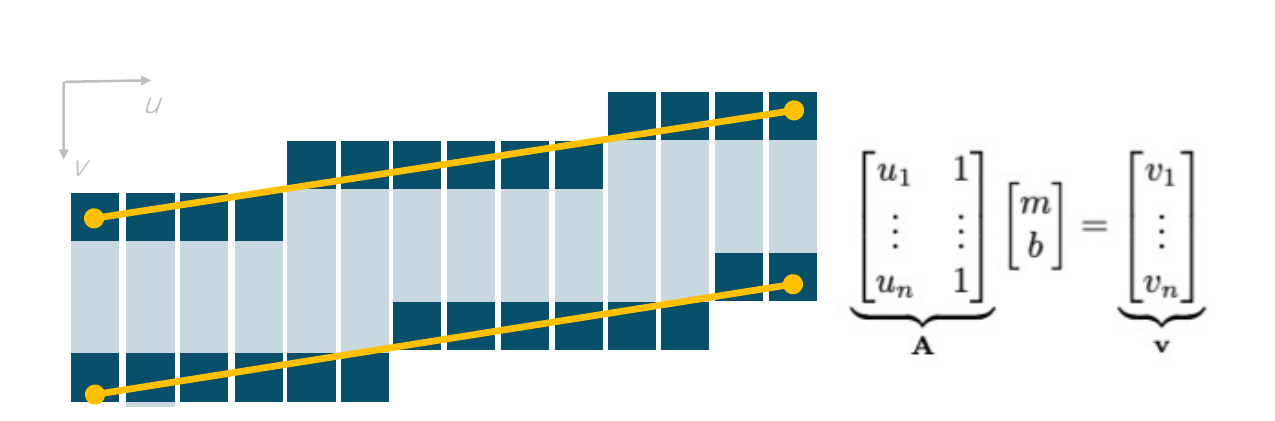}
    \caption{Boundary detection (Step C2) for a horizontal shelf structure. From the edge pixel coordinates $(u_1,v_1),...,(u_n, v_n)$, the boundary line segment derived from an ordinary least squares fit is $v_i = mu_i + b$, where $[m, b]^T = \left({\bf A}^T{\bf A} \right)^{-1}{\bf A}^T{\bf v}$. 
    }
    \label{fig:bd}
\end{figure}

The goal of this step is to detect the relevant structural elements to be mapped. Often objects in built environments can be characterized by planer surfaces and represented by a sparse set of connected points. The output of this step is the 2D point coordinates and semantic label used for defining connections. A block diagram of this process is shown in Figure \ref{fig:detection}, with sample outputs for the shelf facing view. 
\begin{enumerate}
    \item Segment structures. Images are processed by an semantic segmentation neural network pre-trained to identify pixel regions associated with structures of interest: horizontal shelves, vertical shelves (from shelf-facing images) and lights (from ceiling-facing images). A confidence threshold is fixed as a parameter to balance false alarm and missed detections. Four types of errors occur in this stage for the shelf-facing view. First, a structure is not detected because its confidence value is below the threshold. Second, the opposite happens and multiple overlapping masks are given for the same structure. Third, in the length dimension of the shelf structures, the segmentation masks are sometimes incomplete. Fourth, in the width dimension, the mask sometimes exceeds the extent of the shelf face because the network misinterprets the structure edge as being part of the face. As we will discuss, each of these impairments affect the downstream processing steps and eventual performance. 
    \item Detect boundaries. Given the segmentation mask of shelf structure, we identify boundary pixels for each edge and compute best-matched line segments according to an ordinary least-squares fit method, as shown in Figure \ref{fig:bd}.   
    \item Remove outliers. Errors in the segmentation network lead to different types of false detections. We introduce heuristic checks for overlapping structures (keep only the longest one), structures whose edges are askew (remove those whose extended edge intersections lie within the image plane extended by 20\% in each dimension), and vertical structures that are too short (remove those that are shorter than $5/8$ of the vertical image dimension). At this point, the vertical elements can be tracked over the image sequence to determine the total number of sections. 
   \item Determine boundary intersections. Vertical boundary pairs are extended to the image frame edges, and horizontal boundary pairs are extended to the next detected vertical pair. The coordinates of the intersection points are calculated and given a label for the section and "side" (e.g., left or right side of a section). 
 
\end{enumerate}
For the ceiling facing views, a similar process is used to extract the center points for the detected lights. If strip lighting is used, the endpoints would serve as the detected features. 

\subsection{Track structures}
At output of of the detection process (Step C), we have the 2D structure points for each of the images in a view sequence, including labels to indicate the section. The goal of the tracking process (Step D) is to determine the number of horizontal shelves per section, assign correspondences of the intersection points across images, and complete the labels for all points. 

\begin{figure}
    \centering
    \includegraphics[trim=240 220 0 150, clip, width=1.5\linewidth]{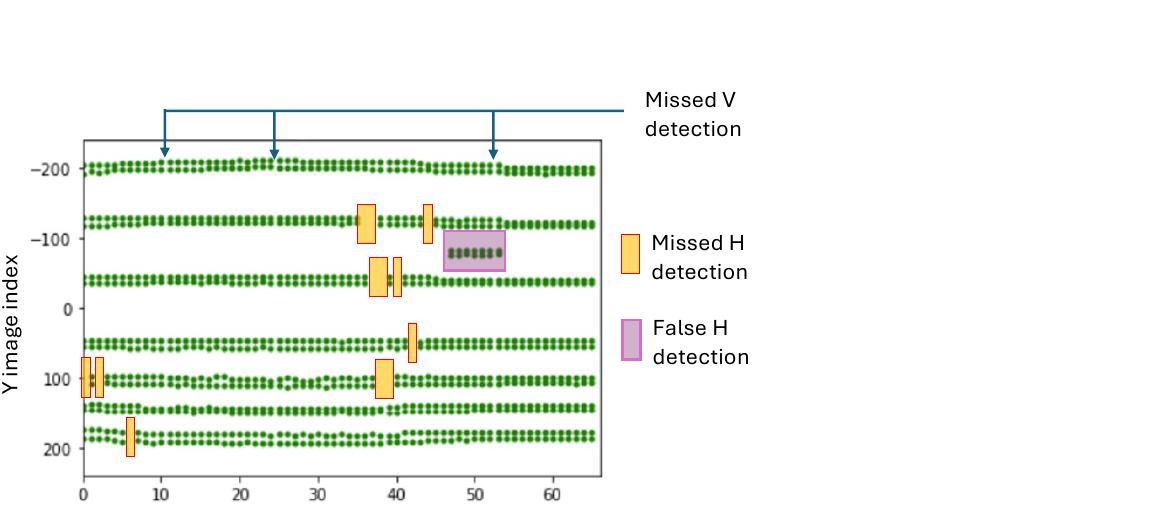}
    \caption{Vertical image coordinates of detected horizontal element points for a section. 
    }
    \label{fig:points2}
\end{figure}

From the sequence of images, we determine those in which a given section's left and right vertical elements are visible. We count the number of horizontal elements by clustering the image coordinates of the left points and distinguishing the groups with a kernel density estimation (KDE) algorithm. For a set of 65 images for a given section, Figure \ref{fig:points2} shows the vertical image coordinates of detected horizontal element points, corresponding to the left intersection points. There is some minor variation in the coordinates due to the camera drift as it moves down the aisle. There are some missing H detections, missing V detection (indicated by blanking), and a set of false H detections (which are rejected).  

\subsection{Constrained SfM}
Given a set of feature points $(S_j)$ tracked across a sequence of $N$ 2D images for a given aisle, the goal of SfM is to estimate the 3D coordinates of the points and the 6DoF camera pose (3D rotation $R_i$ and 3D translation $t_i$) in the world (aisle) frame. This is solved by minimizing the sum error between the point observations $x_{i,j}$ and reprojected points $P(R_i,t_i,S_j)$. We impose real-world geometric constraints on the coordinates to avoid convergence to a spatially meaningless solution. 

For the shelf face case shown in Figure \ref{fig:intersection}, we can write an SfM problem statement, constraining the points to lie on a 2D Manhattan grid:
\begin{align*}
    \min_{R_i, t_i, S_j} &\sum_{i,j} w_{i,j} \, \| x_{i,j} - P(R_i, t_i, S_j) \|^2 \text{ , subject to:} \quad &&\\
    &
    \left.
    \begin{aligned}
    &S_{i,x} = S_{j,x}, \forall i,j \in C_{s,c}, c\in\{0, \dots ,N^{(c)}_s-1 \}\\
        &S_{i,y} = S_{j,y}, \forall i,j \in R_{s,r}, r\in\{0, \dots ,N^{(r)}_s-1 \} \\
        &S_{i,y} = h_{\text{ref},s},  i=R_{s,0}\\
    \end{aligned}
    \right\} \forall s \\
    &\, S_{j,z} = 0, \forall j,
\end{align*}
where, for each section $s$, the first condition constrains any point belonging to the edge of a given vertical element should be aligned with the same $x$ coordinate value. Similarly, the second condition constrains the $y$ coordinate of points belonging to the edge of a given horizontal element. The third condition is a metric constraint for the height of the bottom edge of the lowest horizontal element for each section. The fourth constraint places all points in the $xy$ plane of the world frame. 

\begin{figure}
    \centering
    \includegraphics[trim=100 150 0 120, clip, width=1.4\linewidth]{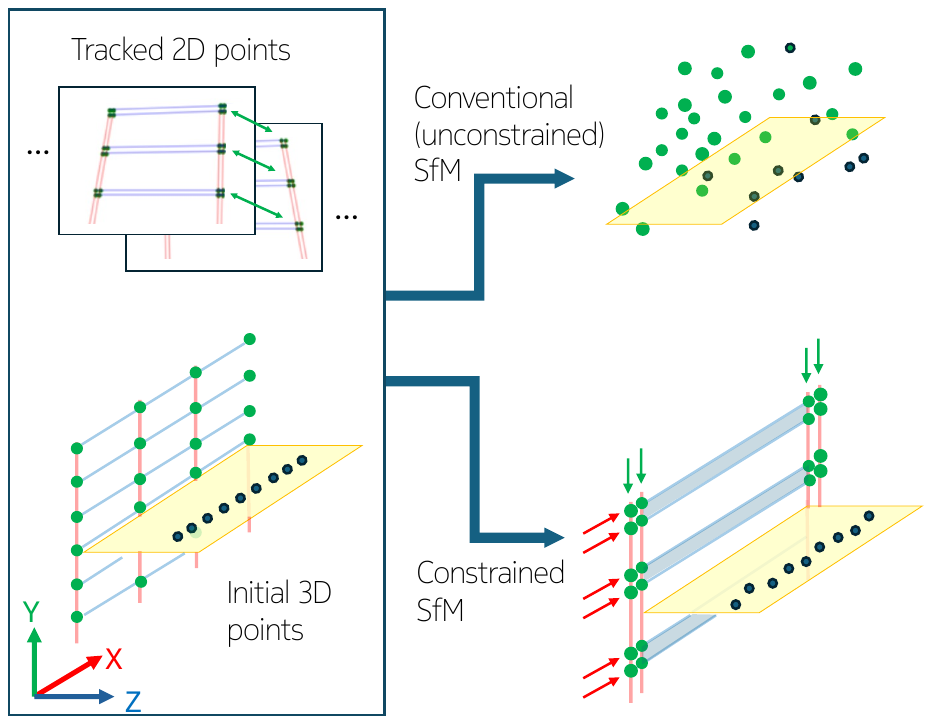}
    \caption{Conventional SfM without structural constraints could lead to a meaningless result. Constrained SfM results in an accurate map.  
    }
    \label{fig:optimization}
\end{figure}

As shown in Figure \ref{fig:optimization}, the shelf point coordinates are initialized to lie on a 2D Manhattan grid on the $xy$ plane with equal spacing. Assuming smooth motion of the camera along the length of the aisle, the initial $X$ coordinate of the camera translation is set to $nL/N$, for image $n = 0,...,N-1$, where L is the approximate length of the aisle. The initial $Y$ and $Z$ coordinates of the camera translations are fixed and set respectively to the approximate camera height and distance from the shelf face. Leveraging the Manhattan grid assumption, the initial 3DoF camera orientation can be computed for each image based on the vanishing points of the largest rectangle formed by the detected horizontal and vertical elements \cite{Wang1991vp}. 




A conventional SfM optimization can be solved using a factor graph framework \cite{Dellaert2017}, where variables are the unknown 3D points and camera poses, and factors are statistical characterizations of the variables based on prior information or measurements. Each factor adds a term to the cost function which the optimizer attempts to minimize. Once an SfM problem has been cast as a factor graph, the optimization can be efficiently solved using software libraries such as GTSAM \cite{gtsam}. 

\begin{figure}
    \centering
    \includegraphics[trim=240 300 0 120, clip, width=1.45\linewidth]{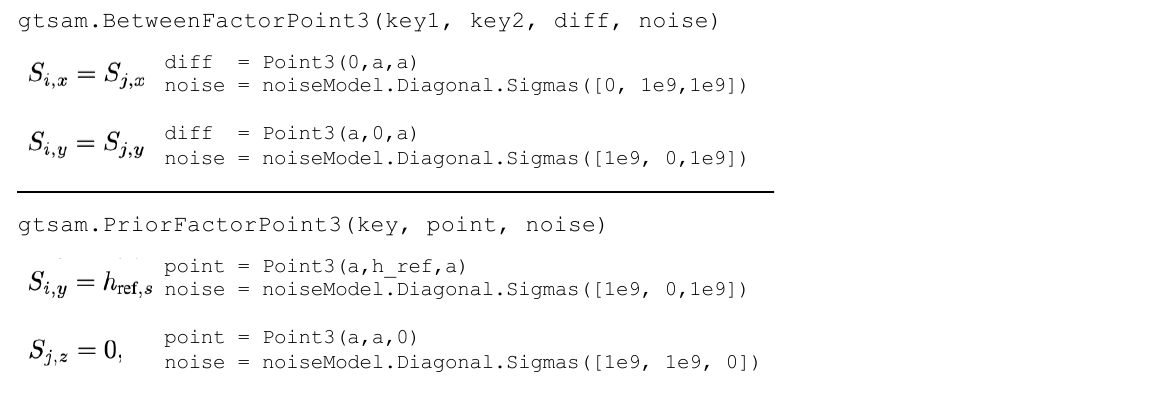}
    \caption{GTSAM code for implementing SfM constraints    }
    \label{fig:code}
\end{figure}

To add the constraints, it is possible to generalize the factor graph framework to incorporate Lagrangian constraints \cite{Bazzana2023} or to introduce custom factors for GTSAM. Instead we simply work with existing GTSAM templated classes \texttt{BetweenFactor} and \texttt{PriorFactor}, shown in Figure \ref{fig:code} The first is used to constrain the translation between two points, and the second constraints a point to a given value. Both constraints are up to a Gaussian noise model. For example, to constrain the $X$ coordinate of points in a column for $S_{i,x},S_{j,x}$, we use the \texttt{BetweenFactor} class and set the difference and noise variance for the $X$ (first) dimension are zero. The difference for the other dimensions are arbitrary (indicated by \texttt{a}, but set in practice to zero), and the variance is set to \texttt{1e9}, indicating no constraint. The \texttt{PriorFactor} is used in a similar fashion.


After the SfM process provides the shelf coordinates and camera poses based on the shelf-facing view, we can turn our attention to the ceiling view and solve a similar SfM problem to determine the light coordinates. Each ceiling view image corresponds to a synchronzed shelf view image. Hence for the ceiling view sequence, we create a new set of camera variables which constrained to have the same position as the corresponding shelf view cameras (which are known and fixed). Geometric constraints can be applied to the lights, such are requiring them to lie on a line parallel to the aisle with a fixed Z if justified by visual observation, and the GSTAM solver can determine the light positions and camera orientations.  


\section{Experiments}
\label{sec:results}
The SAVMap system has been tested on data from a commercial warehouse consisting of 46 shelving rack rows, each about 55m long by 7m high. The width of the aisle between rows is about 2m, and the total shelving footprint is about 8000\,m$^2$. Each row consists of 15 sections, with average width of about 3.7\,m. There is a bridge section in the middle of each row, with 5 to 11 horizontal shelves per section on either side of the bridge. The contents of the shelves is heterogenous and includes boxes of different sizes, wrapped pallets, and loose items. 

\subsection{Preliminaries}
{\bf Panoramic video capture.}  We use a consumer Insta360 X4 panoramic camera mounted at a height of about 3m, less than half the shelf height. Because of the narrowness of the aisle, we required two passes per aisle for the video capture. The cart was pushed at a casual walking speed of about 1m/s, so a single row can be captured in about a minute. Considering the overhead for traveling between aisles, the 46 rows can be captured in an hour. The camera captures a high-resolution (7680 x 3840) panoramic video at 30 frames per second. The video content is compressed as an MP4 file at roughly 1GB per minute for a total of about 60GB. 

{\bf Rendering.} Image sequences of the shelves and ceiling were rendered at VGA resolution (640$\times$480 pixels) at 15 frames per second. For shelf views, a wide angle field of view of about 150$^\circ$ was set by the automated rendering Step B, and the ceiling view field of view was set to 110 degrees. If the camera is 1.5m away from the shelf face, each pixel in the vertical dimension corresponds to about $2.3{\rm cm} = 1500 \tan(150^\circ/2)/(480/2)$. For the ceiling facing images, the focal length was set to 110$^\circ$. 

{\bf Structure detection.} For semantic segmentation, we use the YOLOv8 \cite{yolov8} computer vision model developed for high-efficiency object detection tasks. To provide more robust detection on the light and shelf elements, we fine-tune the baseline model using only a few hundred hand-labeled images drawn from shelf and ceiling-facing sequences. 

{\bf Ground truth.} For each shelving row in the warehouse, a ground truth map from shelf parameters (Horizontal (H) and Vertical (V) widths and gaps) was generated from construction documents and data sheets and verified through manual measurements. The minor expansion of the structure (about 3mm/m in height and 25mm/m in width \cite{rack_guide}\cite{rack_guidelines}) is not accounted for. The light position accuracy is not characterized, so no ground truth is required. 

\subsection{Baseline aisle performance}
\begin{figure}
    \centering
    \includegraphics[trim=170 110 0 100, clip, width=1.5\linewidth]{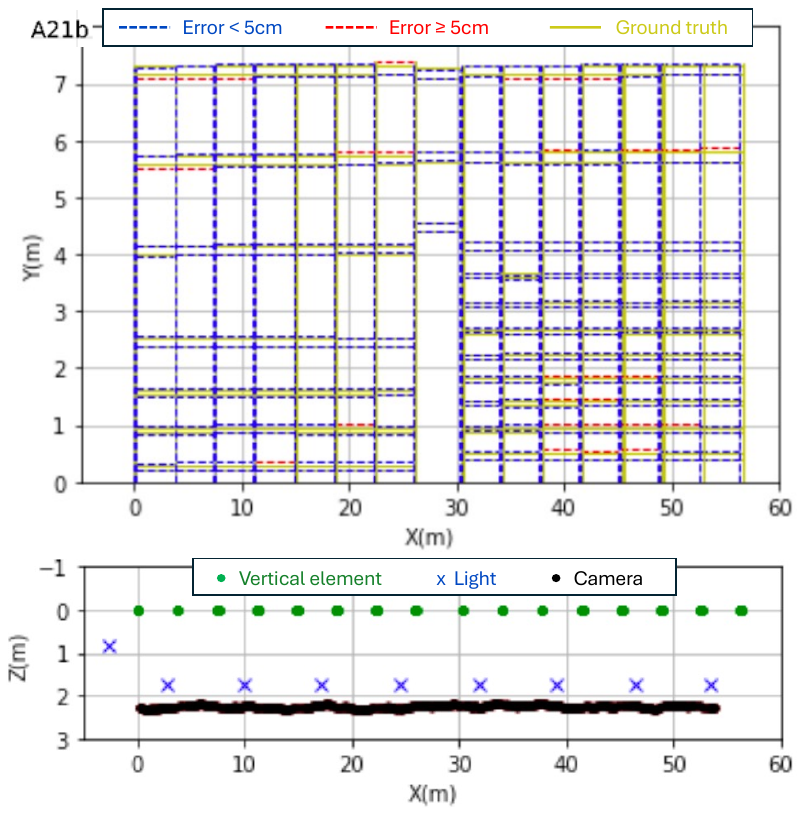}
    \caption{The map of shelves and lights generated by SAVMap for the baseline aisle A21b, shelf-facing (top) and ground-facing (bottom) views. For the shelf view, the mapped shelving (dashed) is superimposed on ground truth.      
    }
    \label{fig:A21F}
\end{figure}

As a baseline, we focus on a single row with dense shelving (up to 11 horizontal elements per section). The map output for the shelves and lights is shown in Figure \ref{fig:A21F} with shelf-facing and ground-facing views. (Note the different scales for the axes.) 

For the shelf view, the mapped shelving is shown as dashed lines on top of the yellow solid ground truth. (Highlighted in red are a minority of horizontal edges with absolute error of more than 5cm.) As a result of the SfM constraints, the shelf elements lie on a plane and conform to a Manhattan grid. The ground view shows the vertical elements, lights, and camera position (starting from about X=0\,m and moving to the right) for 771 images. 

\begin{figure}
    \centering
    \includegraphics[trim=200 180 0 180, clip, width=1.6\linewidth]{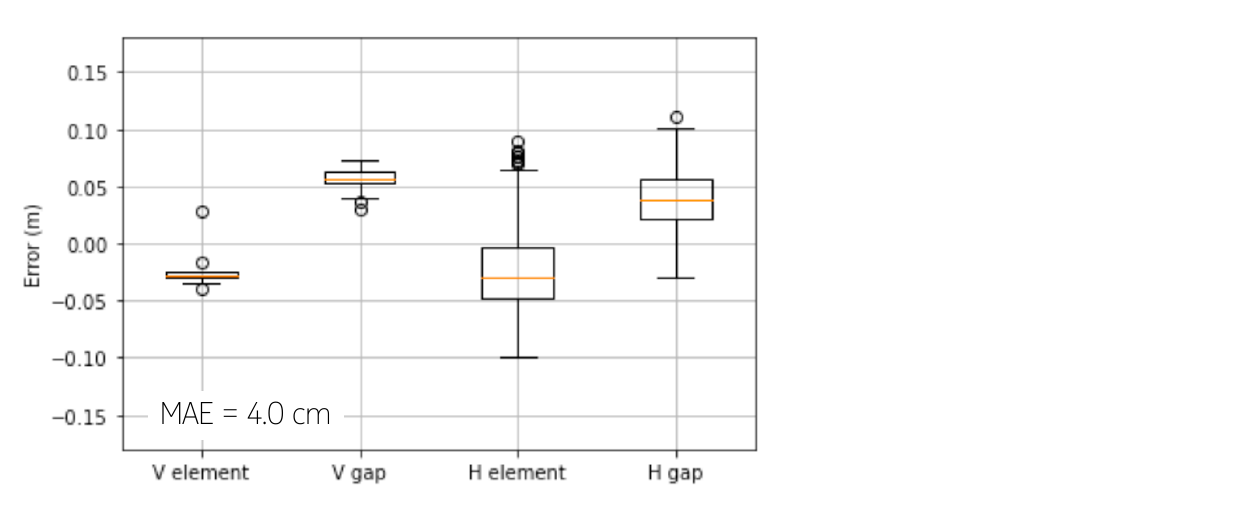}
    \caption{Box plot of map parameter errors compared to ground truth, for baseline aisle A21b in Figure \ref{fig:A21F}. The mean absolute error across the four variable types is 4.0\,cm.     
    }
    \label{fig:A21B_boxplot}
\end{figure}

For each parameter (H or V, element or gap width), we calculate the error $p - \hat p$, where $p$ is the ground truth parameter and $\hat p$ is the estimate. A box plot of the errors, typically in the single-digit centimeter range, is shown in Figure \ref{fig:A21B_boxplot}. The element widths tend to be overestimated (i.e., the mapped element is wider than ground truth) because the computed boundaries at the output of Step C2 tend to lie just outside the structure edges instead of on it. Conversely as a result, the gaps tend to be underestimated. 

For the vertical structures, the absolute gap error slightly exceeds the element error, contributing to an accumulated underestimation of the total aisle width. Hence as seen in Figure \ref{fig:A21F}, the rightmost vertical structure seen at the end of the sequence is to the left of ground truth. For the horizontal structures, the gap and element errors are balanced in magnitude, so the mapped edge coordinates are more accurate but less precise. By aggregating the absolute values of the errors across the 4 classes we can compute the mean absolute error to be 4.0\,cm. 

\subsection{Scaling robustness}
\begin{figure}
    \centering
    \includegraphics[trim=200 180 0 180, clip, width=1.5\linewidth]{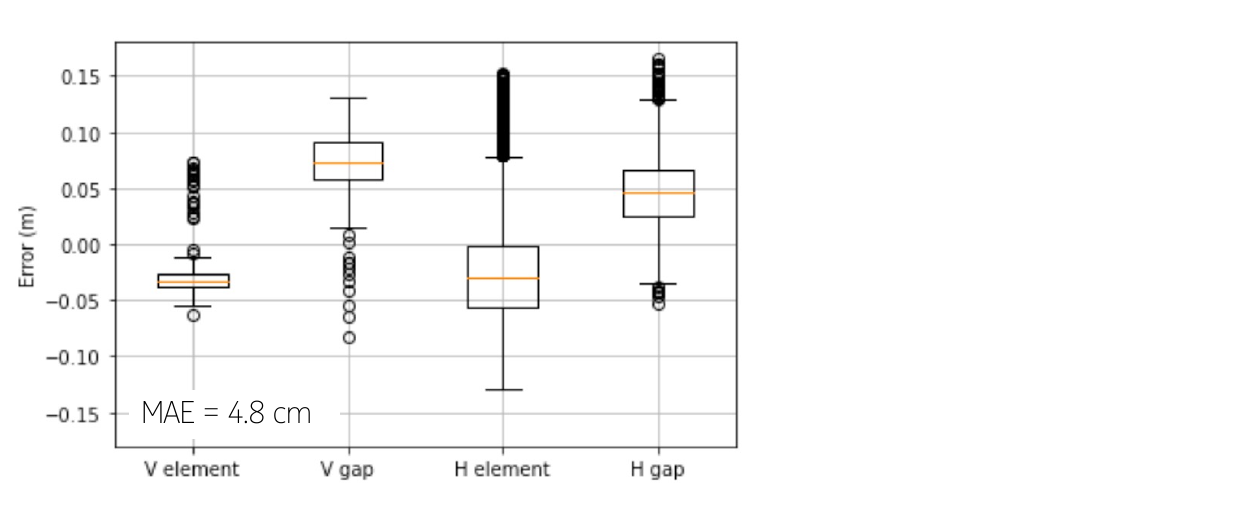}
    \caption{Statistical box plot of shelf parameter estimation errors over all 46 shelving rows. The mean absolute error over the measurements (2890 element width and 2547 gap width) is 4.8\,cm.
    }
    \label{fig:agg_boxplot}
\end{figure}

\begin{figure*}
     \centering
     \includegraphics[trim=180 270 0 80, clip, width=1.55\linewidth]{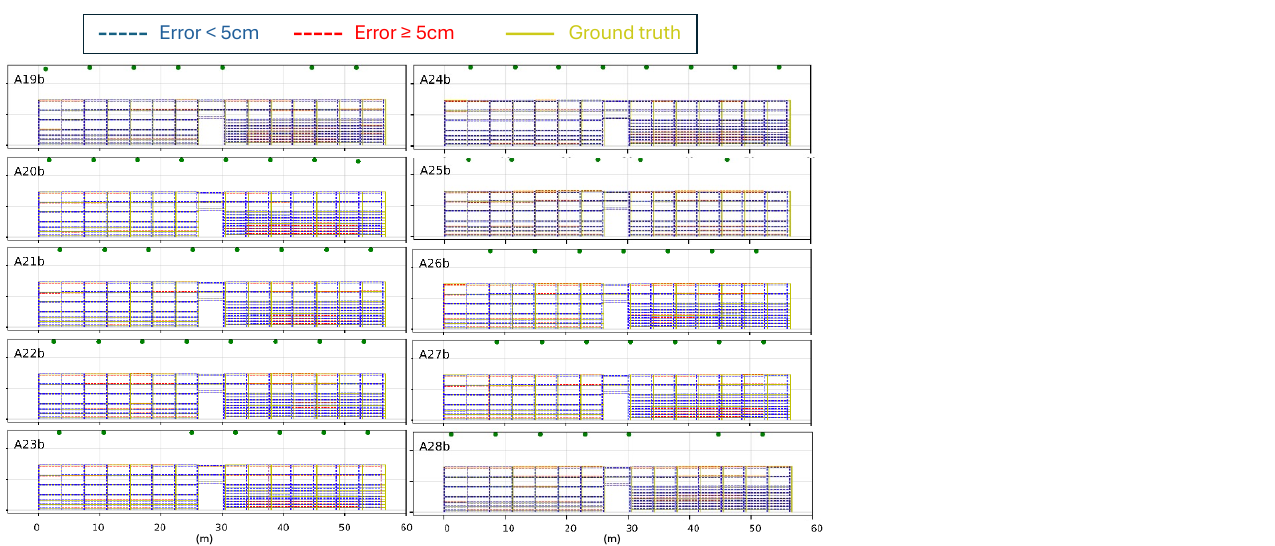}
     \caption{Ten consecutive rows of mapped shelves and lights using SAVMap, including A21b baseline.}
     \label{fig:agg10}
\end{figure*}

We apply the SAVMap process to 46 shelving rows in the warehouse, consisting in total of 736 vertical elements and 4920 horizontal elements and stretching over 2km of horizontal distance. The maps of ten consecutive rows are shown in Figure \ref{fig:agg10}, starting from our baseline (A21b). As we did with the baseline row, the error for each shelf map parameter is computed with respect to ground truth. These errors were aggregated over 46 rows, and the resulting box plot is shown in Figure \ref{fig:agg_boxplot}. The error distributions are wider as result of more outliers across the larger dataset, however the vast majority of errors remain below 10cm. The means (red lines) of each error type remain about the same as the baseline, indicating consistent performance across aisles. The mean absolute error, calculated over the aggregate results is 4.8cm. In total, these results demonstrate that the SAVMap process is robust at scale and has accuracy approaching that of reality-capture devices. 

\subsection{Qualitative comparison}
Using the baseline rendered shelf view image sequence, we use a conventional COLMAP \cite{colmap_sfm}\cite{colmap_mvs} process to generate a dense point cloud. Due to the repetitive shelf structure, the algorithm fails to initialize, even when running in the most computationally intensive exhaustive matching mode. 

\begin{figure*}
     \centering
     \includegraphics[trim=185 270 0 290, clip, width=1.4\linewidth]{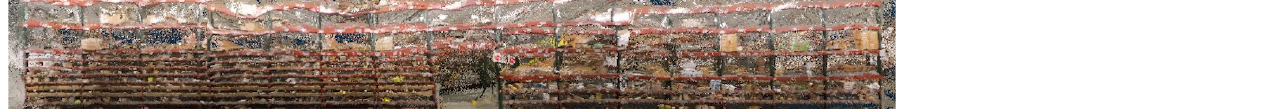}
     \caption{VGGT output from A21b baseline input images.} 
     \label{fig:vggt_wide}
\end{figure*}

We also process the baseline sequence using a state-of-the-art VGGT \cite{wang2025vggt} network for generalized 3D scene understanding. Because of the large memory requirements, the multiple sections were each rendered separately and manually stitched to align the bottom shelves. As seen in Figure \ref{fig:vggt_wide}, the reconstructed aisle is reasonable in some areas but severely distorted in others. Compared to SAVMap, the VGGT representations are not usable for localization and lack semantic meaning. 

\vspace{-0.5cm}

\section{Conclusions}
\label{sec:conclusions}


SAVMap is a low-cost technique for creating accurate, semantic 3D maps of warehouse infrastructure at scale using panoramic image input and leveraging real-world geometric constraints on the structures. This approach was validated in a operational warehouse containing over 2 kilometers of shelving, where we generated detailed maps for every aisle, using just one hour of panoramic video collection. Collectively, the maps achieve a mean absolute error less than 5\,cm versus ground truth, demonstrating the capability to capture large-scale infrastructure efficiently.

Unlike general image-based reconstruction techniques, our specialized mapping approach offers lower cost, reduced computational complexity, and greater robustness in visually repetitive or cluttered environments. The resulting sparse maps can be used for localization, for example, using a similar front end based on semantic segmentation \cite{huang2025savloc}.

While we demonstrated SAVMap on warehouse shelves and lights, it could be extended to map other objects such as signs, labels, boxes, pallets, and containers. The pipeline would need to be augmented to segment and process the new object classes. As we mentioned earlier, fine-tuning the semantic segmentation network for new objects requires training with only a few hundred labeled images, and the remaining processing should be generalizable in a straightforward manner. 

We have implicitly assumed that our proposal is designed for offline generation an initial environment map. As such, the complexity of the processing was not considered. However, the process is actually very lightweight as a result of the YOLOv8's efficiency and processing of the sparse point sets. We have implemented the model on a mobile compute platform and have demonstrated real-time image segmentation up to 5 frames per second. Hence SAVMap could potentially be implemented for real-time, device-based mapping of dynamic environments. 

\section*{Acknowledgment}
The authors would like to thank the following colleagues for their suggestions, guidance, and support: Nilish Suriyachchi, Mouhyemen Khan, Lars Andersson, Amine Houyou, Paul Wilford, Michael Baldwin. Prasanth Ananth, Paul Heitlinger.





\bibliographystyle{IEEEtran}
\bibliography{bibfile}

\end{document}